\documentclass[letterpaper, 10 pt, journal]{IEEEtran}  

\usepackage{amsmath,graphicx}                                                      
\usepackage{amsfonts}
\usepackage{color}

\title{\LARGE \bf
A Versatile Approach to Evaluating and Testing Automated Vehicles based on Kernel Methods
}

\author{Zhiyuan Huang$^{1}$, Yaohui Guo$^{2}$, Henry Lam$^{3}$, and Ding~Zhao$^{2,4,*}$
\thanks{*This work was funded by the Mobility Transformation Center at the University of Michigan with grant No. N021552.}
\thanks{$^{1}$Department of Industrial and Operations Engineering, University of Michigan, Ann Arbor, MI 48109, USA.}
\thanks{$^{2}$Robotics Institute, University of Michigan, Ann Arbor, MI 48109, USA.}
\thanks{$^{3}$Department of Industrial Engineering and Operations Research, Columbia University, New York, NY 10027, USA.}
\thanks{$^{4}$Department of Mechanical Engineering, University of Michigan, Ann Arbor, MI 48109, USA.}
\thanks{$^{*}$Corresponding author. E-mail: zhaoding@umich.edu (Z. D.)}
}

\begin{document}

\maketitle
\thispagestyle{empty}
\pagestyle{empty}

\begin{abstract}


Evaluation and validation of complicated control systems are crucial to guarantee usability and safety. Usually, failure happens in some very rarely encountered situations, but once triggered, the consequence is disastrous. Accelerated Evaluation is a methodology that efficiently tests those rarely-occurring yet critical failures via smartly-sampled test cases. The distribution used in sampling is pivotal to the performance of the method, but building a suitable distribution requires case-by-case analysis. This paper proposes a versatile approach for constructing sampling distribution using kernel method. The approach uses statistical learning tools to approximate the critical event sets and constructs distributions based on the unique properties of Gaussian distributions. We applied the method to evaluate the automated vehicles. Numerical experiments show proposed approach can robustly identify the rare failures and significantly reduce the evaluation time.

\end{abstract}


\section{Introduction}


The auto companies have been competing to get their automated vehicles (AVs) ready on road for years, yet there is still none available in the market. Partly, this is due to the challenging task of robustly testing and guaranteeing the safety of an AV before its release. Companies have been trying different methods such as road test~\cite{wardle_2017,newman_2017}, computer simulation test~\cite{madrigal_2017} and human-vehicle interaction test~\cite{petters,toor_warren_2017}, yet providing safety certificate for an AV system is still open for solving~\cite{wardle_2017}. Assisting the endeavors of solving this problem, the U.S Department of Transportation has released a new AV policy: A Vision for Safety 2.0~\cite{james_2017}. This official document standardizes the required safety features of an autonomous vehicle, providing guidance and clearer pathways for the various stakeholders aiming to certify the safety of their AV systems. However, even with this newly published official guideline, the testing standard remains unclear while the AV target release is quickly approaching. Thus, an effective and efficient testing method for an autonomous vehicle is an urgent need under this background.

Traditional vehicle safety tests are based on crash databases collected from crashes or dangerous scenarios, such as the CSD and GIDAS crash databases~\cite{zhao2017trafficnet}. However, the information logged in these databases is limited so that it is difficult to reconstruct and analyze the dangerous scenarios. As a result, the Naturalistic-Field Operational Test (N-FOT) has been actively used instead in recent auto safety evaluation. An N-FOT logs naturalistic driving data such as the driver’s maneuver, driving environment, and vehicle dynamics, providing detailed data for reconstructing the desired traffic scenario. Among these N-FOT records are the 100-Car Naturalistic Driving Study~\cite{neale2005overview}, the Safety Pilot Model Deployment (SPMD)~\cite{spmd}, and Strategic Highway Research Program (SHRPII)~\cite{punzo2011assessment}. To evaluate the safety of an AV algorithm, an N-FOT based safety evaluation tests the performance of driving policy under a certain sampled environment drawn from the databases. The recent research includes modeling driver failure into safety evaluation~\cite{Chai2017156,kovaceva2015contributing} and analyzing the contributing factors in dangerous scenarios~\cite{dingus2016driver,bella2014analysis,qu2014safety}. In these researches, the safety is evaluated under certain conditions, from which it is then generalized under various naturalistic driving conditions. While sampling from the N-FOT data is a safe method to imitate the reality, it turns out that such approach is deemed inefficient. According to National Highway Safety Administration~\cite{facts2014data}, the fatality rate is 1.08 per 100 million vehicle miles, which means it requires approximately 100 million miles data to generate a fatal crash in a simulation under natural conditions. This enormous simulation effort is indeed costly and time-consuming. Thus, an accelerated method is necessary to attain an effective and efficient auto safety evaluation.

Accelerated Evaluation (AE) procedure is proposed in~\cite{zhao2017accelerated} to serve this notion of accelerated framework, and is further studied in~\cite{zhao2017acceleratedb,huang2017accelerated,huang2017acceleratedb}. This approach utilizes statistical models to represent the traffic environment and estimates the rate of safety-critical events in the test scenarios. Obtaining an unbiased estimator for these rates is accelerated by the use of Importance Sampling (IS), alleviating the needs for extensive simulation replications. The efficiency of this approach depends primarily on the choice of accelerated distribution, which needs to be constructed based on specific characteristics of the problem of interest.



This paper extends the study of AE by proposing a new scheme for constructing accelerated distribution. The scheme merges supervised learning methods with rare event simulation and utilizes the properties of Gaussian distribution in IS theory. The proposed approach provides an alternative method for constructing accelerated distribution for AV test scenarios under Gaussian Mixture Models (GMM).



The structure of the paper is as the following: Section \ref{sec:AE} reviews the AE procedure and emphasizes the concepts relevant to the proposed approach. Section \ref{sec:approach} presents the base knowledge and shows the procedure of the proposed approach. Section \ref{sec:experiments} showcases the performance of the proposed approach based on two problem instances.


\section{Review of Accelerated Evaluation} \label{sec:AE}

In this section, the AE procedure is reviewed in detail in the following order. Section \ref{sec:AE_procedure} gives the high-level overview of the AE procedure, followed by Section \ref{sec:Problem} in which the general problem setting for the approach is described. To provide adequate knowledge regarding the statistical methods used, the IS theory, that supports the validity of AE procedure, will be concisely presented in Section \ref{sec:IS}.


\subsection{Procedure For Accelerated Evaluation}\label{sec:AE_procedure}

In evaluating the performance of an AV system, the AE procedure suggests to examine the AV under various different test scenarios. The safety level of a vehicle is assessed by the rate of safety-critical events in the test scenarios. In this case, the goal of the AE is modeled as a probability estimation problem (described in section \ref{sec:Problem}). The AE procedure consists of four major steps ~\cite{huang2017accelerated}:


\begin{itemize}
	\item Model the behaviors of the traffic environment represented by $f(x)$ (original distribution) as the major disturbance to the AV using large-scale naturalistic driving data.
	\item Skew the disturbance statistics from $f(x)$ to modified statistics $f^{*}(x)$ (accelerated distribution) to generate more frequent and intense interactions between AVs and the traffic environment.
	\item Conduct “accelerated tests” with $f^{*}(x)$. 
	\item Use the IS technique to “skew back” the results to understand real-world behavior and safety benefits.
\end{itemize}

Note that the key part in the AE procedure is the use of IS technique, whose theory will be reviewed in Section \ref{sec:IS}.


The proposed approach provides an alternative in constructing the accelerated distribution in Step 2. In the following sections, we focus on the construction of accelerated distribution. 

\subsection{Evaluation Problem}\label{sec:Problem}
The AE evaluates testing vehicles by individually estimating the rate of safety-critical events under various different test scenarios. Here, we show the setting of the problem and the notations that we use in the following sections.

In a test scenario, we assume that the uncertain environment traffic is represented by a vector $X \in \mathcal{X}$ and is regarded as a stochastic model. Note that $\mathcal{X}$ is the domain of $X$. The distribution model $f(X)$ for $X$ is estimated from a sufficiently large set of naturalistic driving data. The safety-critical events occurs with certain values of $X$. We represent the set as $\varepsilon=\{X| X \text{ leads to a safety-critical event}\}$ and use an indicator function \begin{equation} \label{eq:indicator}
	I_\varepsilon(x)=\begin{cases} 1 & x \in \varepsilon\\
0 & otherwise.\end{cases}
\end{equation} to denote the response of a certain vector $x$. $I_\varepsilon(x)$ is usually complicated, but can be evaluated by on-track experiments or computer simulations.

Generally, the safety-critical events are very rare (smaller than $10^{-6}$). For this reason, using crude Monte Carlo method to estimate the objective is time-consuming due to large sample size needed to observe a critical event, and even much larger samples to achieve statistical confidence. Therefore, we need to improve the efficiency of the evaluation.


\subsection{Importance Sampling}\label{sec:IS}
IS ~\cite{asmussen2007stochastic,ross2013simulation} is a technique to reduce the variance in simulation, while providing unbiased estimation. This technique is crucial to the efficiency and unbiasedness of the AE.

Consider a random vector $X\in \mathcal{X}$ with distribution $f(X)$ and a rare event set $\varepsilon \subset \Omega$ on sample space $\Omega$. Our goal is to estimate the probability of the rare event \begin{equation}
{P}(X \in \varepsilon)=E_f[I_\varepsilon(X)]=\int x f(x) dx,
\end{equation} where the event indicator function is defined as \eqref{eq:indicator}.

Given samples $X_1,...,X_N$ independently and identically generated from $f(x)$, the crude Monte Carlo computes the sample mean of $I_\varepsilon(x)$ \begin{equation}
	\hat{P}(X \in \varepsilon) = \frac{1}{N} \sum_{n=1}^N I_\varepsilon(X_n).
\end{equation}

The IS technique is derived from a change of sampling distribution as the following~\cite{asmussen2007stochastic}:\begin{equation}
	E[I_\varepsilon(X)]=\int I_\varepsilon(x) f(x) dx = \int I_\varepsilon(x) \frac{f(x)}{f^*(x)} f^*(x) dx,
\end{equation}
where $f^*$ is a distribution density function that has the same support with $f$.

Note that 
\begin{equation}
	E^*[I_\varepsilon(x) \frac{f(x)}{f^*(x)} ] = \int I_\varepsilon(x) \frac{f(x)}{f^*(x)} f^*(x) dx,
\end{equation}
where $E^*$ denotes the expectation with regard to $f^*$. Therefore by taking the sample mean of the expectation, \begin{equation}\label{eq:is_estimator}
	\hat{P}(X \in \varepsilon) = \frac{1}{N} \sum_{n=1}^N I_\varepsilon(X_n) \frac{f(x)}{f^*(x)}f(x)dx \end{equation}
gives an unbiased estimator of the above expectation~\cite{asmussen2007stochastic}, where $X_i$'s are generated from $f^*$. 

By appropriately selecting $f^*$, the evaluation procedure obtains an estimation with smaller variance. We refer to $f^*$ as the accelerated distribution in our procedure. The construction of a ``good'' accelerated distribution $f^*$ is discussed in section \ref{sec:approach}.

\section{Constructing Accelerated Distribution via Classification Methods} \label{sec:approach}
\begin{figure}[t]
	\centering
	\includegraphics[width=\linewidth]{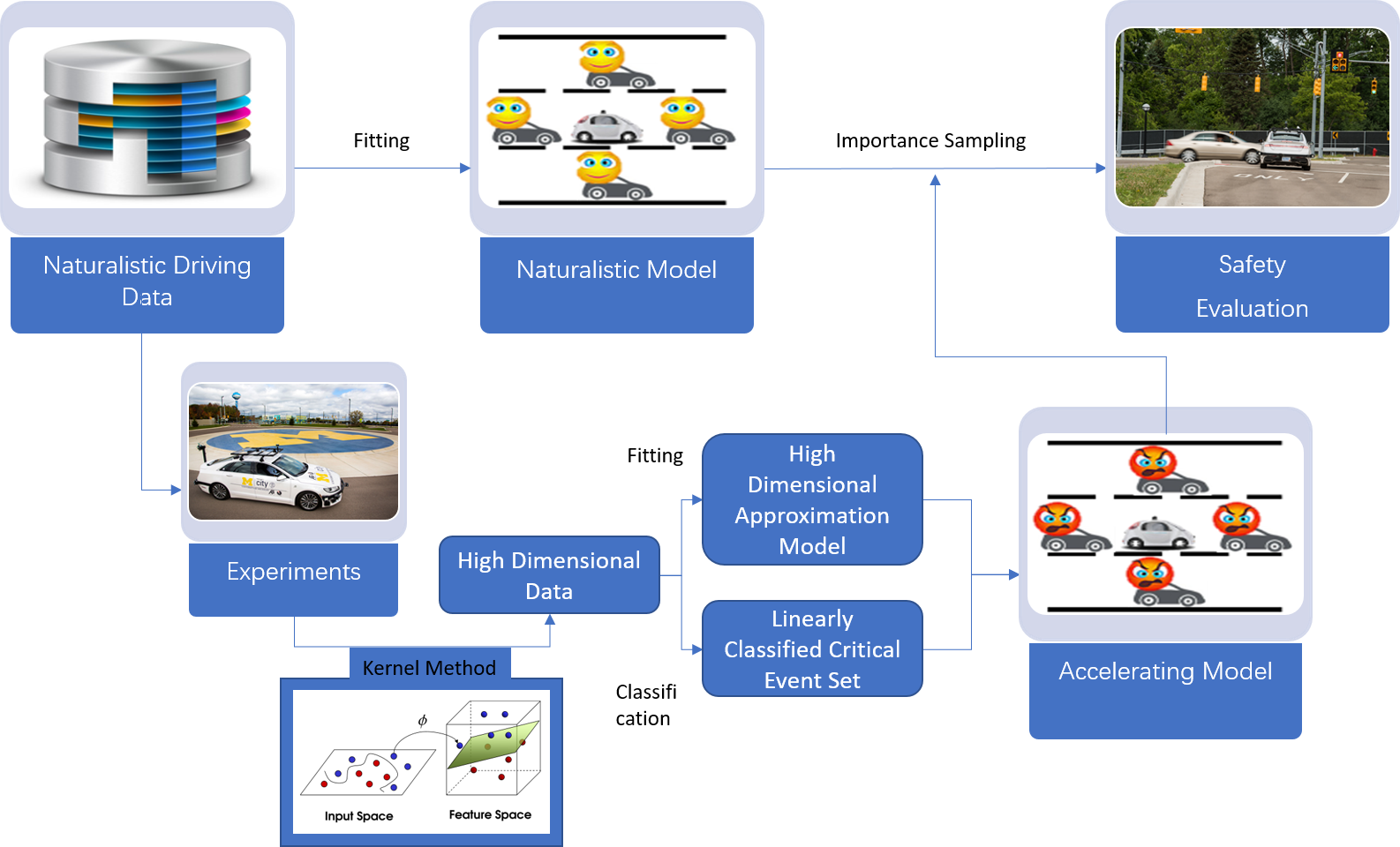}
	\caption{The procedure of the proposed approach.}
	\label{fig:procedure}
\end{figure}
As discussed in~\cite{huang2017acceleratedb}, GMM is a powerful distribution model for representing traffic environment in AE. In this section, we propose a new approach to construct accelerated distributions for test scenarios under GMM setting.

In Section \ref{sec:dominant_point}, we review the existing strategies for constructing efficient accelerated distribution for GMM. Section \ref{sec:construction} shows the key idea of our approach. Section \ref{sec:senquential} presents the procedure of the proposed approach.


\subsection{Known Strategy for Efficient Accelerated Distribution}\label{sec:dominant_point}
When the random vector $X$ follows Gaussian distribution with mean $\mu$ and covariance matrix $\Sigma$ and the rare event set $\varepsilon$ satisfies the convexity assumption, there is a simple scheme that obtains an efficient accelerated distribution~\cite{sadowsky1996monte,sadowsky1990large}.

For a convex rare event set $\varepsilon$, we define the dominating point of $\varepsilon$ on $\phi(x;\mu,\Sigma)$ to be \begin{equation}\label{eq:dom}
	a^*=\arg \max_{a\in \varepsilon} \phi(a;\mu,\Sigma),
\end{equation}
where $\phi(x;\mu,\Sigma)$ is the density function for Gaussian distribution with mean $\mu$ and covariance matrix $\Sigma$. By shifting the mean $\mu$ of the Gaussian distribution to $a^*$, we obtain an accelerated distribution that provides an efficient estimator for $P(x \in \varepsilon)$~\cite{sadowsky1990large}.

\begin{figure}[t]
	\centering
	\includegraphics[width=\linewidth]{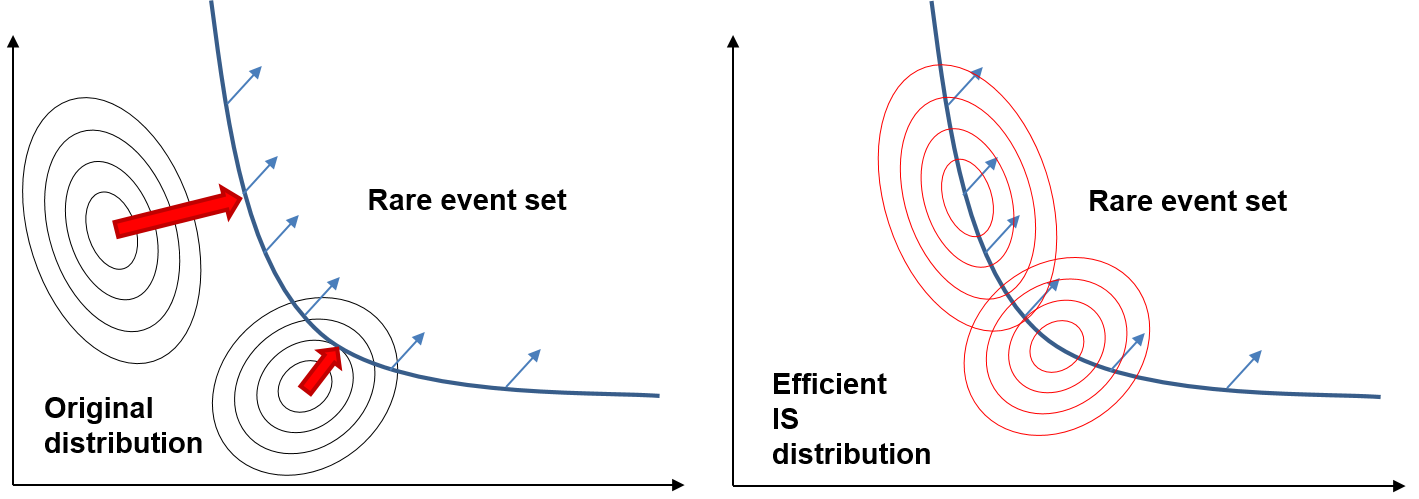}
	\caption{Scheme of constructing accelerated distribution for Gaussian Mixture Model with convex rare event set.}
	\label{fig:stgy1}
\end{figure}

Now we consider the GMM with the density \begin{equation}
	f(x)=\sum_{i=1}^{k}p_i \phi(x;\mu_i,\Sigma_i).
	\label{eq:gmm_density}
\end{equation}
where $k$ is the number of mixture components, $p_i$ is the proportion of the $i$th component, $\mu_i$ and $\Sigma_i$ are the mean and covariance for the $i$th component. We use $a^*_i$ to denote the dominating point with regard to the $i$th component and critical event set $\varepsilon$. An efficient accelerated distribution is then given by:\begin{equation}\label{eq:gmm_acc}
	f^*(x)=\sum_{i=1}^{k} p_i \phi(x;a_{i},\Sigma_i).
\end{equation}
Fig. \ref{fig:stgy1} illustrates this scheme.

In practice, the critical event sets for automated vehicle evaluation are generally non-convex. Therefore, it is hard to apply this approach directly. 


\subsection{Constructing Accelerated Distribution via Kernel Method}\label{sec:construction}

\begin{figure}[t]
	\centering
	\includegraphics[width=\linewidth]{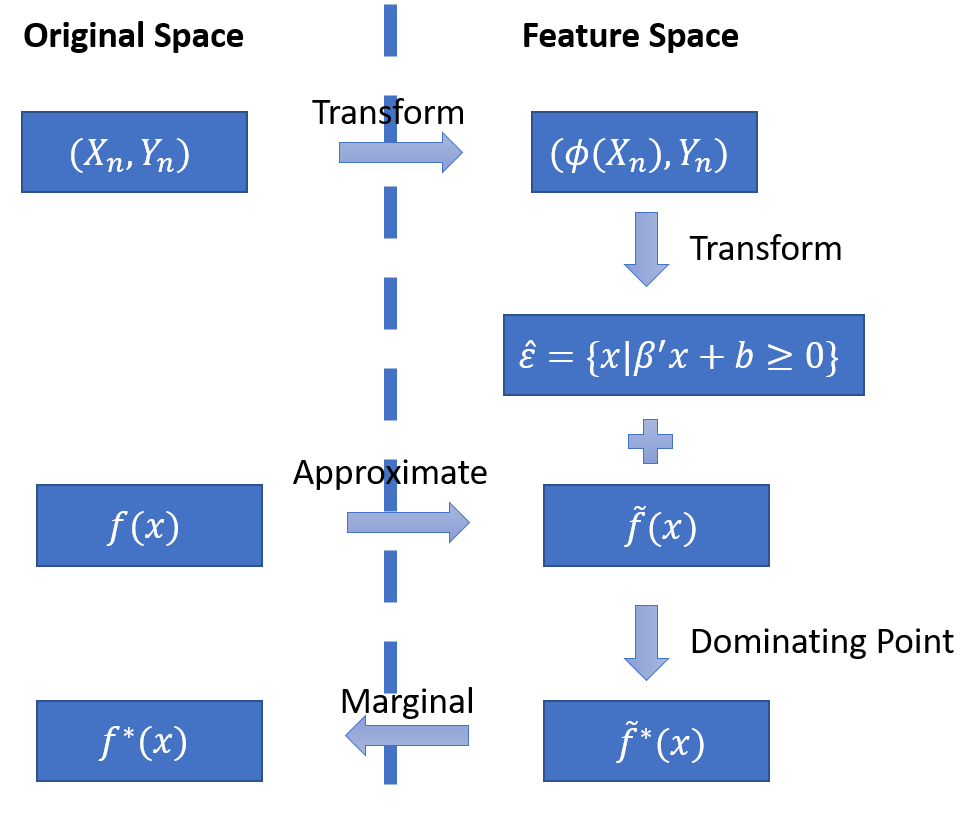}
	\caption{The procedure of constructing an accelerated distribution $f^*$.}
	\label{fig:procedure2}
\end{figure}

We propose a procedure to construct an accelerated distribution for GMM as illustrated in Fig. \ref{fig:procedure2}. The procedure utilizes the scheme in section \ref{sec:dominant_point} and uses kernel method to transform the data on a higher dimensional space. The Classification is used to obtain the linear boundaries for the critical event set.

The following intuition motivates our approach: Suppose we have a feature function (see Appendix B) $\phi:\mathcal{X} \rightarrow \mathbb{R}^m$, where $\mathcal{X} \subset \mathbb{R}^n$ is the domain for $X$ and $n<m$. We assume that there is a subspace on the feature space that is a linear transformation of the original space. (For instance, the feature function for polynomial kernel $\phi(x)=(x^2,\sqrt{2}x,1)$ satisfies the assumption.) We further assume that on the feature space, the critical event set has a linear boundary $\beta'x+b=0$ and $\tilde{f}$ is a Gaussian mixture distribution that well fit the the model on the feature space. Thus, the scheme in section \ref{sec:dominant_point} is directly applicable. Let $\tilde{f}^*$ be the accelerated distribution constructed on the feature space - we note that it is still a GMM. If the feature space contains the a linear transformation of the original space, we can easily obtain an accelerated distribution $f^*$ on the original space by linearly transforming the marginal distribution of $\tilde{f}^*$. Since the marginal of a Gaussian distribution is still a Gaussian distribution~\cite{papoulis2002probability}, $f^*$ is still a Gaussian mixture distribution.

To carry out the procedure described above, we need to find a feature function $\phi(x)$, the boundary $\beta'x+b=0$ and the model $\tilde{f}$. The selection of $\phi(x)$ depends on the shape of the critical event set, which is problem dependent. To obtain $\beta'x+b=0$, we use classification methods, e.g. support vector machine (SVM), to learn for the linear boundaries. The learned critical event set is represented by $\varepsilon=\{x|\beta'x+b\geq 0\}$.

If $f$ is a Gaussian distribution, the true distribution on the feature space is generally not Gaussian. In the proposed procedure, we use a GMM $\tilde{f}$ to approximate the true distribution on the feature space.

\subsection{A Sequential Procedure for the Accelerated Evaluation}\label{sec:senquential}

To summarize the procedure of the proposed approach in section \ref{sec:construction}, we present the step-by-step description and elaborated discussion how to apply the approach in a sequential procedure.

Suppose we have a GMM $f(x)$ for random vector $X$ and we are interested in $E[I_\varepsilon(X)]$. The procedure is as follows:
\begin{enumerate}
	\item Construct a training set with $n$ samples, $\{(X_i,Y_i)\}_{i=1}^{n}$, where $Y_i=I_\varepsilon(X_i)$. We suggest to select $X_i$'s using space filling designs. \label{step:update}
    
	\item Select a feature function $\phi(x)$ and transform the training set as $\{(\phi(X_i),Y_i)\}_{i=1}^{n}$. We suggest to use the feature function for polynomial kernel and use a small parameter $d$ ($d=2$ or $d=3$).
    
    \item Use a classification method with linear boundary on the training set $\{(\phi(X_i),Y_i)\}_{i=1}^{n}$ and obtain the linear boundary $\beta'x+b=0$. We use SVM (see Appendix A) in the experiments.
    
	\item Generate samples $\{X_j\}_{j=1}^{m}$ from the model $f$ and transform the samples as $\{\phi(X_j)\}_{j=1}^{m}$.
    
	\item Select $K$ as the number of mixture components. Use GMM with $K$ mixtures to fit the samples $\{\phi(X_j)\}_{j=1}^{m}$ and obtain the approximation model $\tilde{f}$. The discussion about the selection of $K$ refers to section \ref{sec:implement}
    
    \item For each component in $\tilde{f}$, find the dominating point using \eqref{eq:dom} and construct $\tilde{f}^*$ using \eqref{eq:gmm_acc}.
    
    \item Adjust the marginal distribution of $\tilde{f}^*$ and obtain $f^*$.
    
    \item Generate sample $X_i$'s from the distribution $f^*$ and use \eqref{eq:is_estimator} to estimate the objective. \label{step:sim}
    
    \item (Additional step) As new returns $I_\varepsilon(X_i)$ of $X_i$'s are obtained in step \ref{step:sim}, update $\{(X_i,Y_i)\}_{i=1}^{n}$ and go back to step \ref{step:update}.\label{step:seq}
\end{enumerate}

Note that if we use step \ref{step:seq}, the procedure sequentially update the training set and the information of the critical event set. When the experiment $I_\varepsilon(x)$ is very time-consuming and expensive, we suggest to start with small sample size and add step \ref{step:seq} in the procedure. In the experiments in section \ref{sec:experiments}, we do not use step \ref{step:seq}.

\section{Numerical Experiments}\label{sec:experiments}
In this section, we present simulation results to illustrate the procedure of the proposed approach and to show the applicability of the approach on AV evaluation. In section \ref{sec:toy}, we apply the proposed method on a simple problem to provide an intuition of the mechanism. Section \ref{sec:lane_change} shows the performance of the method in an AV test case.

\subsection{Simple Example Problem}\label{sec:toy}

We use a simple example to discuss the implementation of the proposed method. The example is set to be a low dimensional problem for illustration purposes. Despite being simple, the problem setting maintains some similarity with AV test scenarios.

Assume we have a 2-dimensional random vector $X\in \mathcal{X}$, where $\mathcal{X}\in \mathbb{R}$. Here we have $\mathcal{X}=\{(x,y)|0\leq x \leq 5,0\leq y \leq 5\}$. Suppose the distribution model $f(X)$ of $X$ is known and we have $f(X)$ to be a multivariate Gaussian distribution with mean $(1,1)$ and identity covariance matrix $I_2$. The critical event set is defined as $\varepsilon  =\{(x,y)| x^2+y^2 \leq 0.2^2  \text{ or } (x-5)^2+(y-5)^2 \leq 1.5^2 \text{ or } (x-3)^2+(y-5)^2 \leq 0.7^2 \text{ or } (x-5)^2+(y-3)^2 \leq 0.5^2 \}$. The indicator function $I_\varepsilon(x)={\bf 1}(x \in \varepsilon)$ shows whether a vector $x$ is in the critical event set. Our objective is to estimate the probability of the critical event, i.e. $P(X \in \varepsilon)=E[I_\varepsilon(x)]$.

\begin{figure}[t]
	\centering
	\includegraphics[width=\linewidth]{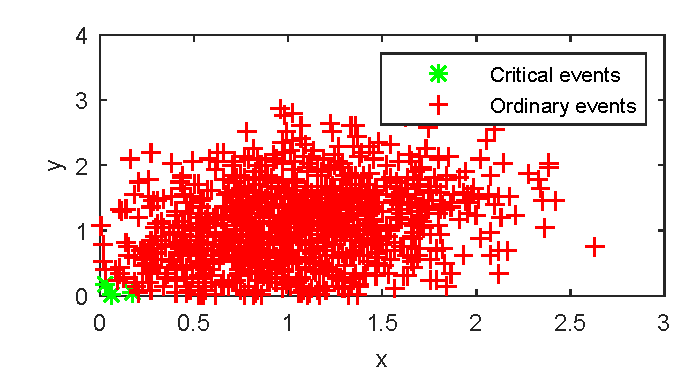}
	\caption{1,000 samples from the distribution model of $X$ with critical events labeled.}
	\label{fig:toy_dist}
\end{figure}

Fig. \ref{fig:toy_dist} shows 1,000 samples generated from the distribution model of $X$. We could observe that distribution concentrates on the area with relatively small value on both axes (consider that the upper bound for both coordinates is 5). 

\begin{figure}[t]
	\centering
	\includegraphics[width=\linewidth]{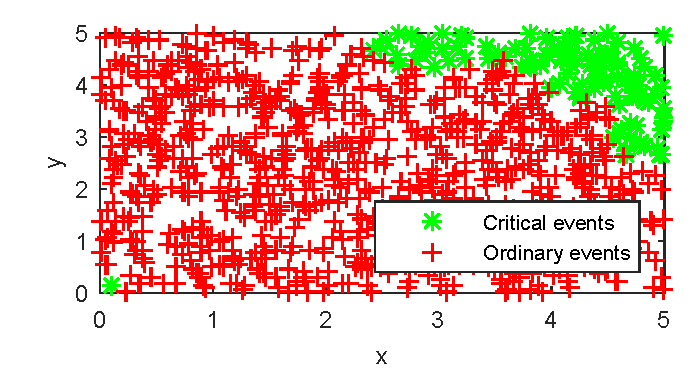}
	\caption{1,000 samples from uniform distribution on the domain of $X$ with critical events labeled.}
	\label{fig:toy_explore}
\end{figure}

Now, we start to explore the domain of the $X$ and to learn the critical event set. We use 1,000 samples randomly generated using a uniform distribution on the domain and label these samples with regard to the critical events. Fig. \ref{fig:toy_explore} shows the samples and the return of the indicator function $I_\varepsilon(x)$. Let us denote these samples as $X_n$ and the returns as $Y_n$. It is obvious that the two types of events cannot be linearly classified and the critical event set is obviously not convex. Therefore if we want to construct an accelerated distribution for estimating the objective probability, we cannot directly use the scheme mentioned in section \ref{sec:dominant_point}.  

\begin{figure}[t]
	\centering
	\includegraphics[width=\linewidth]{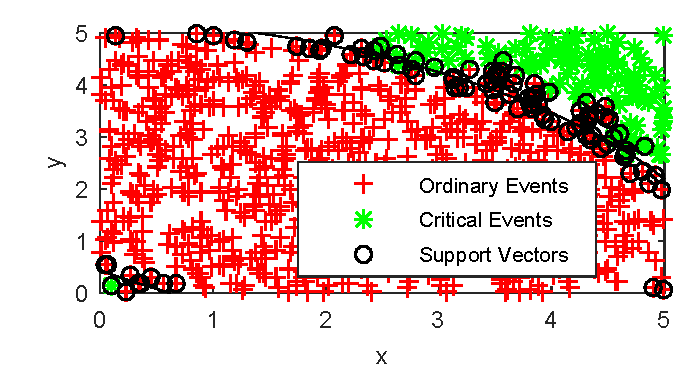}
	\caption{The classification boundary obtained from linear support vector machine on the original space.}
	\label{fig:toy_svm}
\end{figure}

We follow the procedure described in section \ref{sec:senquential} to construct an accelerated distribution for the problem. Here we use a polynomial kernel which maps vectors $X=(x,y)$ in the original space to the feature space through the transformation $\phi(X)=\phi(x,y)=(x^2,y^2,xy,x,y)$. We then use linear SVM to classify the transformed samples $\phi(X_n)$ with regard to the returns $Y_n$. A linear boundary on the feature space is obtained as $\beta'x+b=0$. In Fig. \ref{fig:toy_svm}, we plot this boundary on the original space. We should note that the classification captures the property of the critical set. 

Next, we want to approximate the distribution model on the feature space. We generate extra 20,000 samples from the original distribution $f(X)$, and transform the new generated samples with $\phi(X)$. To investigate the effect of the accuracy of the approximation model, we use two cases with different number of mixture for comparison, i.e. $K_1=20$, $K_2=3$. We fit Gaussian mixture distribution with $K_1$ and $K_2$ components on the transformed samples respectively. For each component in the mixture distribution, we find the dominant point and construct a sampling distribution on the feature space. Finally, we use the marginal distribution on the original space as the accelerated distribution.

\begin{figure}[t]
	\centering
	\includegraphics[width=\linewidth]{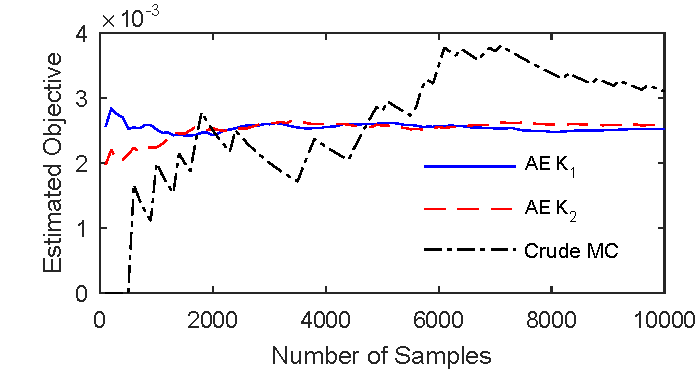}
	\caption{The estimation of the probability $P(X \in \varepsilon)$.}
	\label{fig:toy_est}
\end{figure}

\begin{figure}[t]
	\centering
	\includegraphics[width=\linewidth]{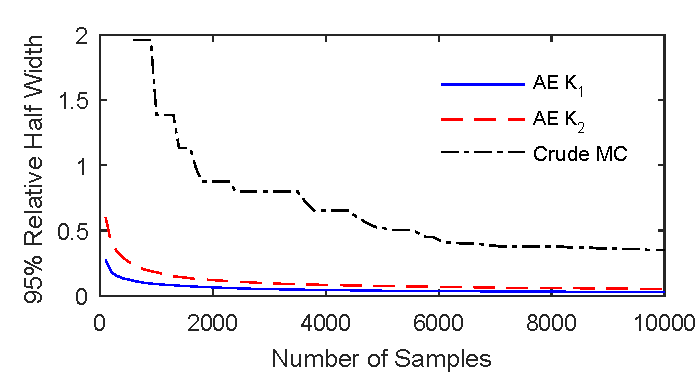}
	\caption{The relative half width with 95\% confidence of the estimation for $P(X \in \varepsilon)$.}
	\label{fig:toy_rhw}
\end{figure}

The simulation performance is shown in Fig. \ref{fig:toy_est} and \ref{fig:toy_rhw}. In the figures, we use ``AE $K_i$'' to denote the proposed approach with $K_i$ components in the approximation mixture distribution and ``Crude MC'' to denote the crude Monte Carlo approach. 

In Fig. \ref{fig:toy_est}, we could observe that the proposed approach attains stability within 2,000 samples. The crude Monte Carlo is obviously oscillating. This shows that the proposed approach provides better estimation using smaller sample size. This argument is further confirmed by Fig. \ref{fig:toy_rhw}. 

In Fig. \ref{fig:toy_rhw}, we plot the 95\% relative half-width, which is given by\begin{equation}
w=\alpha_{0.95} \frac{\hat{\sigma}}{\sqrt{N}\hat{P}} ,
\end{equation}
where $\alpha_{0.95} $ denotes the 0.95 quantile of the normal distribution, $N$ denotes the number of samples used in the estimation, $\hat{\sigma}$ denotes the standard deviation of the sampled objective, and $\hat{P}$ denotes the estimation of the objective.

The 95\% relative half-width for the crude Monte Carlo is much larger than the proposed approaches. To reach the same level of 95\% relative half-width as the proposed approaches, the crude Monte Carlo requires roughly 100 times more samples.

From the performance of this simple example, we shall note that the proposed approach provides good accelerated distributions. By comparing the performance with $K_1$ and $K_2$, we conclude that a more accurate approximation model on the feature space leads to a more efficient accelerated distribution.

\subsection{Lane Change Scenario}\label{sec:lane_change}

\begin{figure}[t]
	\centering
	\includegraphics[width=\linewidth]{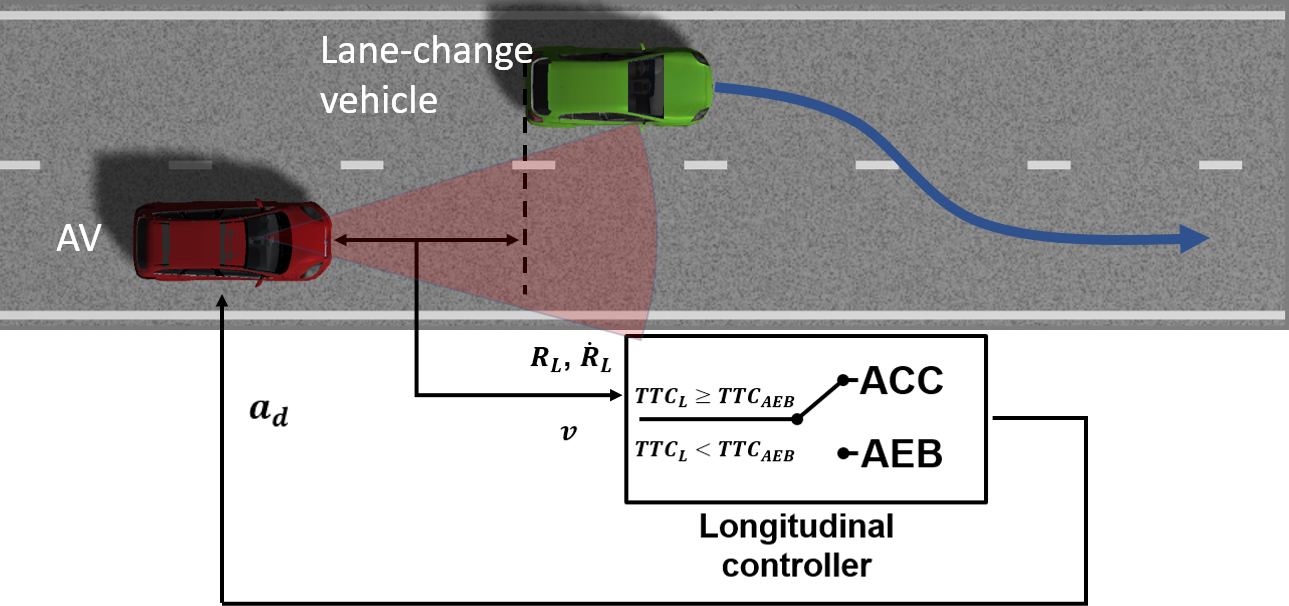}
	\caption{The setting of the lane change scenario and the related parameters.}
	\label{fig:lane_change}
\end{figure}

The lane change model for AE is proposed by~\cite{zhao2017accelerated}. The setting is illustrated in Fig. \ref{fig:lane_change}, which shows that a leading human driving vehicle is conducting a lane change in front of an automated vehicle. Our objective is to evaluate the rate of safety-critical events in this scenario, where safety-critical events includes conflict, collision, injury, etc. In this example, we are interested in the rate of collision during the lane change procedure. We assume that the initial status of the two vehicles is captured by a vector $X=(v,\dot{R},R^{-1}) \in \mathcal{X}$, which consists of three important parameters: $v$, the velocity of the leading vehicle, $\dot{R}$, the relative range of the two vehicles, and $R^{-1}$, the inverse of the range between the two vehicles. We assume that $X$ is stochastic and the distribution of $X$, $f(X)$, is a known GMM. Given $X$, the lane change procedure is supposed to be deterministic with regard to the testing vehicle. We use $\varepsilon \subset \mathcal{X}$ to denote the critical event set and our objective is represented by $P(X \in \varepsilon)$.

In this example, we use the vehicle control model shown in Fig. \ref{fig:lane_change} as the testing vehicle. The model consists of Adaptive Cruise Control (ACC) and Autonomous Emergency Braking (AEB) \cite{ulsoy2012automotive}. (In the figure, $TTC$ is defined as $TTC=R/\dot{R}$. $TTC_{AEB}$ is a threshold that triggers the AEB system.)


We apply the proposed approach to obtain accelerated distributions. We construct a training set with $20,000$ samples by using a grid on the domain ($v$ and $\dot{R}$ are bounded; $R^{-1}$ is unbounded, we sample between the minimum and maximum data in a naturalistic driving data). In this case, we use a polynomial kernel with feature function $\phi(X)=\phi(v,\dot{R},R^{-1})=(v^2,\dot{R}^2,(R^{-1})^2,v\dot{R},vR^{-1},\dot{R}R^{-1},v,\dot{R},R^{-1})$ to map the model onto a higher dimensional space. Again, we use different value, $K_1=8$ and $K_2=20$, for the number of mixtures of the approximation model in the feature space.

\begin{figure}[t]
	\centering
	\includegraphics[width=\linewidth]{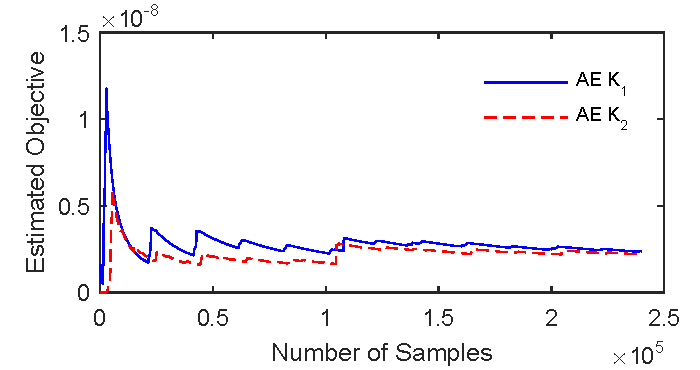}
	\caption{The estimation of the probability $P(X \in \varepsilon)$.}
	\label{fig:lane_est}
\end{figure}

Fig. \ref{fig:lane_est} presents the estimation results of the objective, $P(X \in \varepsilon)$. We observe that the estimation converges as we increase the number of samples. In this case, using different value for the number of mixtures for the approximation model does not result in significant difference in the resulting estimation.

To illustrate the efficiency of the proposed method, we will compare the performance of the proposed method with crude Monte Carlo. Since the probability we are estimating is very small, running crude Monte Carlo would require a huge computing time. Here we use the probability we estimated to approximate the number of samples required for the crude Monte Carlo. Using the formula for the standard deviation of the crude Monte Carlo estimation $\hat{P}(x \in \varepsilon)$ by \begin{equation}
std(\hat{P}(x \in \varepsilon))=\sqrt{\frac{\hat{P}(x \in \varepsilon)(1-\hat{P}(x \in \varepsilon))}{n}},
\end{equation} 
we found that the crude Monte Carlo method requires roughly $6 \times 10^4$ times more samples to reach the sample confidence level as the proposed approach. 

\begin{table}[t]
\centering
\caption{Comparison between the proposed approach and crude Monte Carlo.}
\label{table:compare}
\begin{tabular}{|l|l|l|}
\hline
            & Standard Deviation            & Required Samples                \\ \hline
Proposed AE & $2 \times 10^{-7}$ & $2 \times 10^5$    \\ \hline
Crude MC    & $5 \times 10^{-5}$ & $1.2 \times 10^{10}$ \\ \hline
\end{tabular}
\end{table}

Table \ref{table:compare} summarizes the comparison of the proposed AE with crude Monte Carlo. Column 2 shows the approximated standard deviation of the estimators. Column 3 presents the approximated number of the required samples to obtain an estimation with a 95\% confidence relative half-width less than 0.4. Note that the proposed AE uses an extra of $2 \times 10^4$ samples to learn the critical event set and construct the accelerated distribution. These samples are much smaller than the scale of the required samples for estimation.

\subsection{Discussion on Implementation}\label{sec:implement}

In the implementation of the proposed approach, there are some open choices in the procedure. We know from the experiment  that the efficiency of the approach is largely influenced by these choices. Here, we discuss how they are related to the efficiency of the method.

To obtain an efficient accelerated distribution, we want to achieve two tasks: a) the classification must learn the property of the critical event set, at least a rough bound needs to be found; b) the approximation on the feature space needs to roughly capture the shape of the transformed distribution. We discuss these two tasks separately.

In the classification step, we need to construct a train set and choose a kernel function. Since the collection of train data requires test experiment, the size of the training set need to be balanced. In practice, to select a reasonable number requires prior knowledge of the test scenario. When such knowledge is not available, we suggest to start with a small size of data that includes some critical events and then follow the sequential procedure described in section \ref{sec:senquential}. Given the number of data in the training set, we can use a deterministic design (e.g. use a grid) or a random design (e.g. uniformly sample), as long as samples fill the domain well. 

The kernel functions to select need to have an explicit feature function. We suggest to use polynomial kernel. A feature space with higher dimension would achieve better accuracy for the critical event learning, however, to approximate the distribution model on the higher dimension is generally harder. For this reason, we want the dimension of the feature space to be as small as possible.

For approximating the distribution model on the feature space, we need to choose the number of mixture components. Although a larger number of components always provides a better fitting, the efficiency of the constructed accelerated distribution might not always improve. We suggest to use smaller number of components for less computing efforts. For high dimensional feature space, regularization should be applied in fitting the approximation model to make the fitting algorithm converge in fewer iterations. The selection of regularization parameter is suggested to choose through cross validation~\cite{bishop2006pattern}.

\addtolength{\textheight}{-12cm}   



\section*{APPENDIX}
\setcounter{subsection}{0}
\subsection{Classification Methods for Critical Event Set Learning}\label{sec:svm}
Here we review SVM as an example of classification methods that suits the proposed approach. Please refer to~\cite{bishop2006pattern} for more details.

SVM is a popular algorithm for classification and regression. Here we briefly introduce the SVM with hard margin for binary classification. Denote the training dataset as $(x_i,t_i)$, $  i=1,2,3,...,n$, where $x_i$ is the feature vector and $t_i \in \{ +1,-1\}$ is the label; suppose the data ${x_i}$ is linear separable. The SVM returns the linear classifier of the form:
$$ y(x) = \beta'x+b$$
where $\beta'$ is the coefficient of the hyper plane in the feature space; $b$ is the bias; $y(x_i)>0$ for $t_i=1$; $y(x_i)<0$ for $t_i=-1$. 
SVM wants to find the hyperplane to separate the points while maximizing the minimum distance between the hyperplane and the nearest points $x_i$ from either side. This can be formulated with the constraints as 
    \begin{equation*}
    \begin{aligned}
    &\mathop{\arg\max}_{ \beta,b} \text{   }\{\frac{1}{\lVert \mathbf{\beta} \rVert } \mathop{\arg\min}_{n}{\text{   }t_i(\beta'x_i+b)}\}\\
   &t_i(\beta'x_i+b)>0, \quad i = 1,2,3,...,n
    \end{aligned}
    \end{equation*}
A common solution to SVM is to use the Lagrange Multiplier, which can be found in \cite{bishop2006pattern}.

Note that other classification methods, e.g. logistic regression, also provide a linear boundary and are compatible with kernel method. These methods also fits the proposed approach.

\subsection{Kernel Method for Critical Event Set Learning}\label{sec:kernel}

The kernel is a generalized inner product, computing the inner product of two vectors in a higher dimensional space without explicitly define the feature function. Here we introduce the feature function, and more detail on kernel method can be found in~\cite{murphy2012machine} .The feature function is defined as $\phi:\mathcal{X} \rightarrow \mathbb{R}^m$, where $\mathcal{X} \subset \mathbb{R}^n$ is the domain for input $x_i$, $  i=1,2,3,...,n$, and $n<m$. A benefit of introducing feature space is non-separable data can be separable in some feature spaces. 

For instance, consider two sets on a 2-D plane: $S_1 = \{(x,y) \mid x^2+y^2 < 3\} \subset \mathbb{R}^2 $ and $S_2 = \{(x,y) \mid x^2+y^2 > 3\}\subset \mathbb{R}^2 $. There is no line can separate $S_1$ from $S_2$ since $S_1$ is surrounded by $S_2$. But if we map the point $(x,y)$ on $\mathbb{R}^2$ to $\mathbb{R}^3$ by a feature function $\phi(x,y) = (x,y,x^2+y^2)$, the mapped sets are linearly separable. Denoting the original sets $S_1$, $ S_2$ in the feature space as $\Phi(S_1) = \{\phi(x) \mid x \in S_1 \}$ and $\Phi(S_2) = \{\phi(x) \mid x \in S_2 \}$, then $\Phi(S_1)$ and $\Phi(S_2)$ are linearly separable by the plane $z = 3$ in $\mathbb{R}^3$.




\bibliographystyle{ieeetr}
\bibliography{citation.bib}

\end{document}